\documentclass[journal]{IEEEtran}
\usepackage{microtype}                 
\PassOptionsToPackage{warn}{textcomp}  
\usepackage{textcomp}                  
\usepackage{mathptmx}                  
\usepackage{times}                     
\usepackage{cite}                      
\usepackage{tabu}                      
\usepackage{booktabs}                  
\usepackage{subfig}
\usepackage{times}
\usepackage{epsfig}
\usepackage{graphicx}
\usepackage{amsmath}
\usepackage{amssymb}
\usepackage{algorithm}
\usepackage{color}
\usepackage{algorithmic}
\usepackage{bbm}
\usepackage{multirow}
\usepackage{comment}
\usepackage[table]{xcolor}
\usepackage[switch]{lineno}
\newcommand\norm[1]{\left\lVert#1\right\rVert}
\definecolor{myc1}{gray}{0.8}

\newcommand{\gu}[1]{{\textcolor{black}{#1}}}

\usepackage[breaklinks=true,bookmarks=false]{hyperref}

\ifCLASSINFOpdf
\else
\fi
\begin{document}
%
\title{Example-based Color Transfer with Gaussian Mixture Modeling}
%
%
%

\author{Chunzhi Gu,
        Xuequan Lu,
        and~Chao~Zhang
\thanks{Manuscript received XX, xx; revised XX. (Corresponding author: C. Zhang.) }
\thanks{C. Gu and C. Zhang* are with the School of Engineering, University of Fukui, Fukui, Japan (e-mails: gu-cz@monju.fuis.u-fukui.ac.jp; zhang@u-fukui.ac.jp).}
\thanks{X. Lu is with School of Information Techonolgy, Deakin University, Australia (e-mail: xuequan.lu@deakin.edu.au).}
}

\maketitle

\begin{abstract}
Color transfer, which plays a key role in image editing, has attracted noticeable attention recently. It has remained a challenge to date due to various issues such as time-consuming manual adjustments and prior segmentation issues. In this paper, we propose to model color transfer under a probability framework and cast it as a parameter estimation problem. In particular, we relate the transferred image with the example image under the Gaussian Mixture Model (GMM) and regard the transferred image color as the GMM centroids. We employ the Expectation-Maximization (EM) algorithm (E-step and M-step) for optimization. To better preserve gradient information, we introduce a Laplacian based regularization term to the objective function at the M-step which is solved by deriving a gradient descent algorithm. Given the input of a source image and an example image, our method is able to generate continuous color transfer results with increasing EM iterations. Various experiments show that our approach generally outperforms other competitive color transfer methods, both visually and quantitatively. 
\end{abstract}


%
\IEEEpeerreviewmaketitle

\section{Introduction}

\IEEEPARstart{E}{diting} an image by endowing it with the color style of another image (i.e., color transfer) is challenging, because the color style can be significantly affected by various factors such as illumination or saturation. Color transfer, which aims at transforming color from an example image to a source image such that two images appear consistent in color and the transferred result looks natural, is a fundamental problem in image editing and modeling. 

Color transfer techniques based on the source image only \cite{lischinski2006interactive,cohen2006color} often require time-consuming manual adjustments. To reduce the efforts involved in parameterization and subjective judgment, example (i.e., reference, exemplar or target) based color transfer methods have been introduced. The problem can be reduced to interpreting the color of the example image over the source image. The majority of existing example-based methods \cite{tai2005local, xiang2009selective, hristova2015style,tai2007soft,bonneel2013examplebased} require image segmentation as a pre-process to allocate colors to different regions, which is known as local color transfer. Some researches \cite{tai2005local,bonneel2013examplebased} improve the accuracy of segmentation to better capture color variations spatially. Nevertheless, local methods are prone to failure for small regions with insufficient numbers of pixels.  Still, global color transfer algorithms \cite{reinhard2001color,Xiao2006ColorTI,pitie2007automated,pitie2007linear,rabin2014non,ferradans2013regularized} can sometimes generate color that is  inconsistent with the example image, and tend to strike a balance between many kinds of features of an image, which leads to undesirable over-saturation or artifacts \cite{hristova2015style}. As a result, those color transfer methods have limited applicability in the real world.

\begin{figure}[t]
\centering
\setcounter{subfigure}{0}
\subfloat[Source Image]{\includegraphics[width=25mm,scale=0.5]{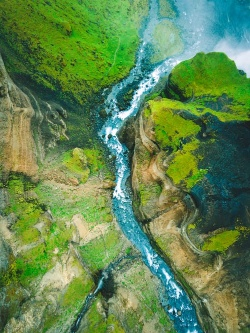}}\hspace{1.0em}%
\subfloat[Example Image]{\includegraphics[width=25mm,scale=0.5]{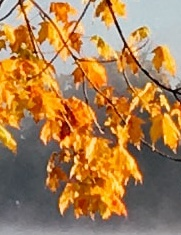}}\hspace{1.0em}%
\subfloat[$q = 3$]{\includegraphics[width=25mm,scale=0.5]{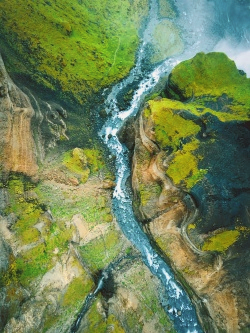}}
\\[0.1ex]

\subfloat[$q = 8$]{\includegraphics[width=25mm,scale=0.2]{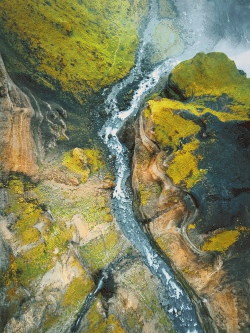}}\hspace{1.0em}%
\subfloat[$q = 15$]{\includegraphics[width=25mm,scale=0.2]{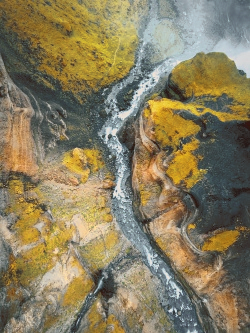}}\hspace{1.0em}%
\subfloat[$q = 35$]{\includegraphics[width=25mm,scale=0.2]{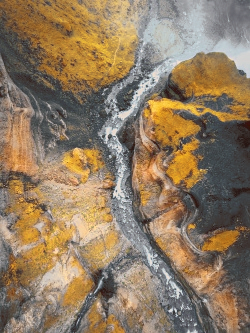}}

\caption{An example of continuous intermediate results produced by our algorithm with (a) and (b) as inputs. (c) $\sim$ (f) are four intermediate color transfer results at different iterations.}
\label{fig:titleFigure}
\end{figure}

In this paper, we propose a novel global color transfer approach, by relating the transferred image of the source image with the example image under the Gaussian Mixture Model (GMM). It does not require any prior segmentation knowledge or results. More specifically, we suppose that the color distribution of the example image follows a GMM, and take the transferred image color (pixels) as the GMM centroids. 
The optimization is achieved via the Expectation-Maximization (EM) algorithm (E-step and M-step) \cite{dempster1977maximum}. To further retain the gradient information of the source image, we introduce a Laplacian regularization term to the objective function in the M-step. This improves color mapping by softly imposing gradient information of the source image to the transferred image, thus alleviating artifacts (e.g., discontinuity in the color transition). Also, we introduce a gradient descent algorithm to solve the objective function in the M-step. 
It is interesting that the proposed approach is able to generate a series of \textit{continuous} color transfer images with the increasing optimization iterations for the EM procedure (Fig. \ref{fig:titleFigure}). To our knowledge, most previous methods can only output a single color transfer result given a pair of images (source image and example image), while our approach enables the flexibility in generating different color transfer results and offer multiple choices for users.

\begin{figure*}[t]
\begin{center}
\includegraphics[width=0.7\linewidth]{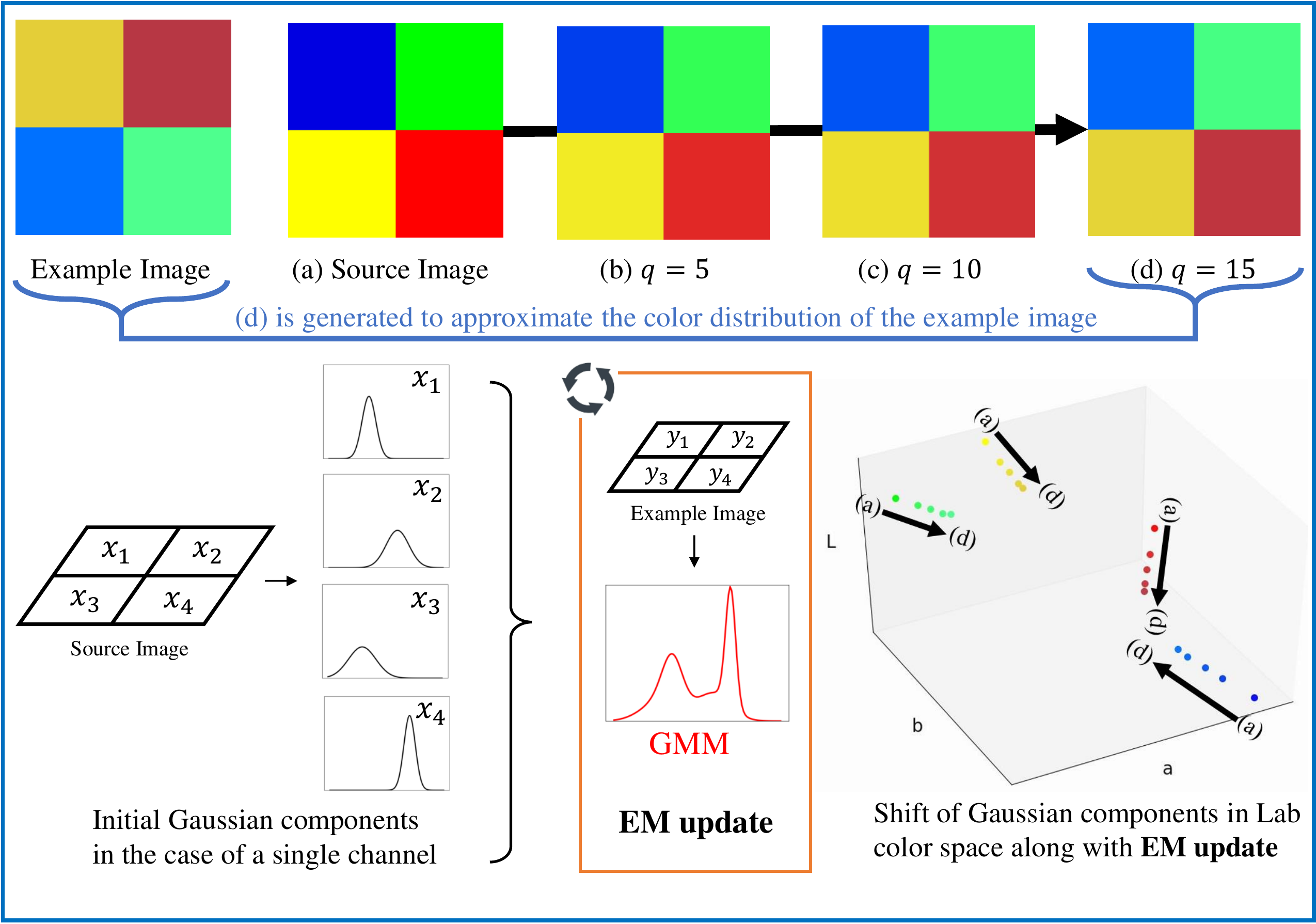}
\end{center}
\caption{Overview of our color transfer approach with images in 2$\times$2 pixels for clarity. Example image and (a) are inputs. (b) $\sim$ (d) are the results after $q$ iterations. Intuitively, the color style of the example image is gradually transferred to the source image, which is modeled by the shift of Gaussian components in the Lab color space under our GMM framework. }
\label{fig:overview}
\end{figure*}

Experiments validate the proposed method and demonstrate that our method generally produces better results than other competitive color transfer techniques. The \textit{main contributions} of this paper are summarized as follows. 
\begin{itemize}
    \item We propose a novel color transfer approach which enables continuous color transfer results given an example image and a source image as input.
    \item We introduce a regularization term to better preserve the gradient information, and derive a gradient descent algorithm to solve the objective function in the M-step.
    \item We conduct extensive experiments and demonstrate that our method generally outperforms other competitive techniques. 
\end{itemize}

\section{Related Work}
In this section, we first review previous example based color transfer techniques. We then introduce the Gaussian mixture model in a few visual computing tasks, especially the color transfer. We finally review some deep learning based color transfer methods.

\subsection{Example based Color Transfer} 
Transferring the color style of an image using an example image
has been extensively studied \gu{\cite{reinhard2001color,pitie2007linear,Xiao2006ColorTI,pitie2007automated,hwang2014color,wu2013content,frigo2014optimal, tai2005local, xiang2009selective, hristova2015style,tai2007soft,xu2005uniform,rabin2014non, ferradans2013regularized, wang2017, nguyen2014illuminant}}. It has a wide range of applications with which users can create a variety of rendering effects by changing the example image. A pioneering work was introduced by Reinhard et al. \cite{reinhard2001color}, in which the color distributions of the source and example images based on the global statistics (i.e., mean and standard deviation) in the uncorrelated $l\alpha\beta$ space are matched with a linear transformation. However, it fails to transfer these dominant statistics when the example image requires non-linear color mapping. Several other approaches \cite{pitie2007linear,Xiao2006ColorTI,pitie2007automated} have been proposed to improve the results later. Xiao et al. \cite{Xiao2006ColorTI} introduced the covariance matrix to better explain the statistics, and freed the restriction in $l\alpha\beta$ space. F. Pitie et al. \cite{pitie2007linear} presented the Monge-Kantorovitch (MK) theory on linear transformation. In some cases, simple linear transformation is limited in color mapping performance. 
F. Pitie et al. \cite{pitie2007automated} designed a non-linear iterative transfer to match pdfs (probability density functions) of two images, followed by a post-processing step to refine the results.

Other efforts have also been made to handle the color transfer problem. 
Hwang et al. \cite{hwang2014color} proposed the probabilistic moving least squares which can preserve gradients of the distribution in mapping colors. The method by Wu et al. \cite{wu2013content} is able to match regions according to the semantic relationship between two images before recolorization. In addition, optimal transportation based approaches \cite{frigo2014optimal, rabin2014non, ferradans2013regularized} have been investigated. Frigo et al. \cite{frigo2014optimal} imposed semantic constraints on the displacement cost of the color mapping function. Ferradans et al.  \cite{ferradans2013regularized} relaxed the mass conservation constraints of the discrete transport and applied it to the color transfer problem. It is followed by Rabin and Papadakis \cite{rabin2014non} which added a spatial regularization term to the transport map to mitigate artifacts.


\subsection{Gaussian Mixture Model (GMM)} 
GMM is a powerful statistical model and has been exploited in some visual computing tasks. Lu et al. \cite{Lu2018} proposed to relate the noisy point cloud and its cleaned version via GMM for point cloud filtering. GMM has also been extended to handle articulated skeleton learning and extraction \cite{lu2018unsupervised,Lu2019}. Gu et al. \cite{gu2019kernel} introduced a GMM based approach for image deblurring. In color transfer, GMM has been mainly exploited for prior image segmentation \cite{tai2005local, xiang2009selective, hristova2015style,tai2007soft,xu2005uniform}. In other words, theses works benefit from the segmentation results achieved by GMM, 
which yields the improvement of color coherence in mapping results. Tai et al. \cite{tai2005local} modeled each segmented region probabilistically with GMM to facilitate soft region boundaries for color
transfer. To further ensure a seamless transition, multiple example images are introduced by Xiang at el. \cite{xiang2009selective}. Hristova et al. \cite{hristova2015style} partitioned images into Gaussian clusters with a pre-judging procedure to identify whether the image is light or color based in nature, in order to flexibly alter the learning data for GMM to improve color transfer effects. However, small regions with insufficient numbers of pixels may be misclassified into other Gaussian components, which weakens its sensitivity of segmentation. By contrast, our method extends GMM to straightforwardly model the pixel-wise color distribution rather than segmentation, which is largely different from the above works and is able to produce natural results.

\subsection{Deep Learning based Color Transfer}
It is worth mentioning that the deep neural networks are often more exploited in the form of ``style'' transfer than color transfer, due to the strong capacity of convolutional neural networks (CNNs) in capturing latent features. Deep features are usually extracted from pre-trained networks to build the relationship between two images (source image and example image) \cite{he2018deep,xiao2019example,he2019progressive,liao2017Visual,luan2017deep}.  In spite of the outstanding performance, deep learning based methods generally require significant computational costs and labor efforts (e.g., long-term training and requirement of GPUs), and a pre-defined size of input images. Unlike our method, given a single image pair as input, none of those methods can produce continuous transfer results. 



\section{Proposed Method}

Fig. \ref{fig:overview} illustrates the overview of our method. We first formulate the color transfer problem under the framework of GMM in Lab color space \gu{(i.e., CIELAB color space)}, and then adopt the EM algorithm \cite{dempster1977maximum} 
to optimize the involved parameters. In other words, we regard color transfer as a parameter estimation problem. We further introduce a gradient regularization term into the objective function at the M-step, for better preservation of gradient information. 

\subsection{The Probabilistic Model}
Given the transferred result $X$ initialized by the source image $S$, and the example image $Y$, we denote the pixel sets $X=\begin{Bmatrix}x_1, \dots, x_m, \dots, x_M\end{Bmatrix}$, $x_m \in \mathbb{R}^3 $, and $Y=\begin{Bmatrix}y_1, \dots, y_k, \dots, y_K\end{Bmatrix}$, $y_k \in \mathbb{R}^3$, respectively. Each pixel holds three channels defined by the Lab color space, which is designed to approximate the human visual perception \cite{robertson1990historical, edge2011method}. 
Our core idea is to model the color distribution by assuming that $Y$ follows a GMM with $X$ as the Gaussian centroids. Thus, the probability dense function for $y_k \in Y$ can be formulated as 
\begin{equation}
\label{eq:eq1}
p(y_k)=\sum_{m=1}^M\frac{1}{M}p(y_k|x_m),
\end{equation}
where $p(y_k|x_m) = \frac{1}{\sqrt{(2 \pi)^{d}|\Sigma_{m}|}} e^{-\frac{(y_k-x_m)^\intercal{\Sigma_{m}}^{-1}(y_k-x_m)}{2}}$ denotes the $m$-th Gaussian component, and $d$ is the dimension of $x_m$ and $y_k$ ($d=3$ due to 3 channels of Lab color space). For the $m$-th Gaussian component, $\Sigma_{m} = \sigma^2_m\boldsymbol{I}$ denotes the diagonal covariance matrix, in which $\boldsymbol{I}$ is the identity matrix, and $\sigma^2_m$ is a scalar whose value varies with $m$. Besides, all the Gaussian components are assigned an equal membership probability $\frac{1}{M}$. 

Notice that we do not set equal covariance values $\sigma^2_m$ for all Gaussian components, in order to allow a more general setting with diverse Gaussians. The centroids of the GMM model are initialized by $X$. The continuous color transfer can be realized by alternately finding the centroids and covariance matrices that best explain the distribution of $Y$. 

\subsection{EM Optimization}
The estimation of centroids \gu{$X$} and covariance matrices \gu{$\Sigma = \begin{Bmatrix}\Sigma_1, \dots, \Sigma_m, \dots, \Sigma_M\end{Bmatrix}$} of the GMM can be achieved by minimizing the following \textit{negative} log-likelihood function \cite{bishop2006pattern,Lu2018},
\begin{equation}
\label{eq:eq2}
E(X,\Sigma)=-\sum_{k=1}^K\log(\frac{1}{M}\sum_{m=1}^Mp(y_k|x_m)).
\end{equation}

The Expectation-Maximization (EM) algorithm \cite{dempster1977maximum} is employed to solve Eq. (\ref{eq:eq2}). There are two steps (E-step and M-step) in the EM algorithm, which are alternately called for multiple iterations to reach decent estimations.

\textbf{E-step.} The posterior probability $p^{old}(x_m|y_k)$ is calculated based on the Bayes' theorem and the parameters updated in the previous iteration. We rewrite $p^{old}(x_m|y_k)$ as $p^{old}_{mk}$ for simplicity,
\begin{equation}
\label{eq:eq3}
p^{old}_{mk} = \frac{ e^{-\frac{(y_k-x_m)^\intercal{\Sigma_{m}}^{-1}(y_k-x_m)}{2}}}{\sum_{m'=1}^M e^{-\frac{(y_k-x_{m'})^\intercal{\Sigma_{m'}}^{-1}(y_k-x_{m'})}{2}}}.
\end{equation}

\textbf{M-step.} Based on the computed posteriors in the E-step, the M-step is to update the involved parameters ($X$ and $\Sigma$), which is equivalent to minimizing the upper bound of Eq. (\ref{eq:eq2}), as shown below
\begin{equation}
\begin{aligned}
\label{eq:eq4}
Q(X,\Sigma) &= -\sum \limits_{k=1}^{K} \sum \limits_{m=1}^{M} p^{old}_{mk} \log \frac{p^{new}(y_k \vert x_m)} { p^{old}(y_k \vert x_m)p^{old}_{mk}}\\
&\propto
\sum \limits_{k=1}^{K}\sum \limits_{m=1}^{M} p^{old}_{mk}\frac{\norm{y_k-x_m}^2}{2\sigma^{2}_{m}} +\sum \limits_{k=1}^{K}\sum \limits_{m=1}^{M}\frac{p^{old}_{mk}}{2}\log \sigma_m^2,
\end{aligned}
\end{equation}
where the superscript ``$new$'' indicates the calculation of the posterior probability with the parameters to be estimated in the current iteration.

\begin{figure}[t]
	\centering
\setcounter{subfigure}{0}
\subfloat[Source image]
{
\includegraphics[width=0.3\linewidth]{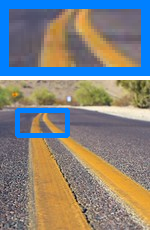}

\label{subfloat:with_regularization_term}
}
\hfill
\subfloat[Without $R(X)$]
{
\includegraphics[width=0.3\linewidth]{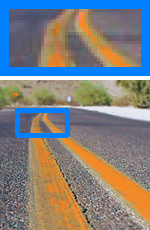}

\label{subfloat:with_regularization_term}
}
\hfill
\subfloat[With $R(X)$]
{
\includegraphics[width=0.3\linewidth]{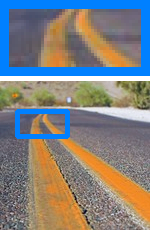}
}
\label{subfloat:with_regularization_term}

\caption{Color transfer results with and without the regularization term $R(X)$. }
\label{fig:regularizationComparison}
\end{figure}

\begin{algorithm}[tb]
\caption{Continuous Color Transfer}
\begin{algorithmic}[1]
\label{alg:algorithm1}
\renewcommand{\algorithmicrequire}{\textbf{Input:}}
\renewcommand{\algorithmicensure}{\textbf{Output:}}
\REQUIRE Source image $S$, example image $Y$.
\ENSURE Result images $\{X^{q}\}, q=1,\dots,q_{max}$
\STATE Initialization: ${\sigma}^2_m = 5$, $\mu \in [0.001,0.005]$, $\alpha=0.1$, $q_{max} = 50$, $q=0$, $X^{0}=S$ \REPEAT
\STATE E-step: update each $p^{old}_{mk}$ by Eq. (\ref{eq:eq3})
\STATE M-step: update each $x_m$ via Eq. (\ref{eq:eq9}), and ${\sigma}^2_m$ by Eq. (\ref{eq:eq10})
\STATE $q=q+1$, save $X^q$
\UNTIL $q_{max}$ is reached
\end{algorithmic} 
\end{algorithm}

\subsection{Regularization Term}
\label{sec:Regularization term}
Eq. (\ref{eq:eq4}) can be regarded as a data term, which can be explained as the data fidelity of $X$ to $Y$. However, this data term only takes the color distribution into account and ignores the spatial information. As illustrated in Fig. \ref{fig:regularizationComparison}(b), 
artifacts could appear (e.g., loss of sharp features and discontinuity in the color transition) without considering spatial information. Inspired by the Laplacian filter \cite{burt1983laplacian}, we introduce a Laplacian regularization term to the objective function Eq. (\ref{eq:eq4}) in the M-step to mitigate this problem. The regularization term $R$ is defined as
\begin{equation}
\label{eq:eq5}
R(X) = \frac{1}{2}\sum_{m=1}^{M}\norm{\Delta X - \Delta S}^2,
\end{equation}
where $\Delta$ is the discrete approximation of Laplacian operator and $S$ is the original source image (i.e., initial $X$ in the first EM iteration). Eq. (\ref{eq:eq5}) can be rewritten as
\begin{equation}
\label{eq:eq6}
R(X) = \frac{1}{2}\sum_{m=1}^{M}\norm{\sum_{m'\in \mathbb{N}_m}w_{m'}\left(x_{m'}-x_m\right) - \Delta S_{m}}^2,
\end{equation}
where $\mathbb{N}_m$ is the set of neighboring pixels within the window of the Laplacian kernel, and $w_{m'}$ is the corresponding coefficient of the discrete Laplacian kernel. 
Each Lab channel is normalized for the distance calculation. Laplacian filter extracts gradient information from an image. This regularization term can help constrain the gradient information of the source image to the transferred image, in a least squares manner. As a result, artifacts can be greatly suppressed and smoother color transition can be obtained, as demonstrated in Fig. \ref{fig:regularizationComparison}(c).

By re-defining Eq. (\ref{eq:eq4}) as $D(X,\Sigma)$, we can rewrite the final objective function in the M-step as
\begin{equation}
\label{eq:eq7}
Q(X,\Sigma) = D(X,\Sigma) + R(X).
\end{equation}

\subsection{Minimization}

Now we explain how to solve $x_m$ and $\Sigma_m$. 
We first take the partial derivation of Eq. (\ref{eq:eq7}) with respect to $x_m$ by assuming the neighbors $\{x_{m'}\}$ are known, 

\begin{equation}
\label{eq:eq8}
\begin{aligned}
\frac{\partial Q(X, \Sigma)}{\partial x_m} &= \frac{1}{{\sigma_m^2}}\sum \limits_{k=1}^{K}p^{old}_{mk}(x_m-y_{k})\\ &-(\sum_{m'\in \mathbb{N}_m}w_{m'})((\sum_{m'\in \mathbb{N}_m}w_{m'}\left(x_{m'}-x_m\right) )- \Delta S_{m}).
\end{aligned}
\end{equation}

We derive a simple gradient descent algorithm to solve $x_m$, based upon Eq. \eqref{eq:eq8}. 
The new $x_m$ 
(i.e., $x_m^{new}$) can be updated as
\begin{equation}
\label{eq:eq9}
\begin{aligned}
x_{m}^{new} =  x_{m} &+ \alpha \frac{\sum \limits_{k=1}^{K}p^{old}_{mk}(y_{k} - x_{m})}{\sum\limits_{k=1}^{K}p^{old}_{mk}}\\ & + \mu ((\sum_{m'\in \mathbb{N}_m}w_{m'}(x_{m'}-x_m) )- \Delta S_{m}),
\end{aligned}
\end{equation}
where $\mu = {\alpha\sigma_m^2\sum_{m'\in \mathbb{N}_m}w_{m'}/\sum_{k=1}^{K}p^{old}_{mk}}$, and $\alpha$ is a hyper-parameter. 
Eq. (\ref{eq:eq9}) is obtained by empirically scaling the gradient of Eq. (\ref{eq:eq8}) with a factor of ${\alpha\sigma_m^2/\sum_{k=1}^{K}p^{old}_{mk}}$, for better control of the gradient descent step. In this work, we take $\mu$ as a controllable hyper-parameter for simplicity. 

After solving $x_m$, we need to update $\Sigma_m$. Since $\Sigma_m = \sigma^2_m\boldsymbol{I}$ is related to the scalar $\sigma^2_m$, we take the partial derivative of Eq. (\ref{eq:eq7}) with respect to $\sigma^2_{m}$. By solving ${\partial Q}/{\partial \sigma^2_{m}}=0$, we can obtain 
\begin{equation}
\label{eq:eq10}
\begin{split}
{{(\sigma}^2_m)^{new} } = (\sum_{k=1}^K p^{old}_{mk} \Vert x_{m}^{new}-y_k\Vert^2 ) /\sum_{k=1}^{K}p^{old}_{mk}.
\end{split}
\end{equation}

We summarize our proposed method in Alg. \ref{alg:algorithm1}. For efficiency, we simply set the number of iterations for Eq. \eqref{eq:eq9} to 1. Our approach is capable of generating continuous color transfer results by increasing the number of EM iterations $q_{max}$. 

\section{Experimental Results}

\begin{table*}[t]
\centering
\caption
{Comparisons of SSIM and PSNR on test images. }
\label{tab:metricsEvaluation}

\begin{tabular}{lccccccccccc}
\hline
\multicolumn{12}{c}{SSIM}                                                                                                                                                                                       \\
\multicolumn{1}{l}{} & butterfly      & parrot         & house          & tulip          & bus            & river          & sea            & flower1        & flower2        & temple         & \textbf{ave.}        \\ \hline
Reinhard et al. \cite{reinhard2001color}      & 0.88           & 0.58           & 0.47           & 0.70           & 0.82           & 0.85           & 0.84           & \textbf{0.86}  & 0.64           & 0.68           & 0.73           \\
Xiao et al. \cite{Xiao2006ColorTI}          & 0.86           & 0.64           & 0.59           & 0.71           & 0.74           & 0.88           & 0.78           & 0.69           & 0.45           & 0.71           & 0.70           \\
Petit et al. \cite{pitie2007linear}          & 0.91           & 0.56           & 0.56           & 0.76           & \textbf{0.88}  & \textbf{0.90}  & \textbf{0.88}  & 0.85           & 0.64           & 0.71           & 0.76           \\
Petit et al. \cite{pitie2007automated}        & 0.86           & 0.70           & 0.60           & 0.71           & 0.79           & \textbf{0.90}  & 0.83           & 0.81           & 0.62           & 0.71           & 0.75           \\
Ferradans et al. \cite{ferradans2013regularized}    & 0.84           & 0.42           & 0.52           & 0.58           & 0.68           & 0.57           & 0.68           & 0.74           & 0.56           & 0.73           & 0.63           \\
Rabin \& Papadakis \cite{rabin2014non}  & 0.76           & 0.57           & 0.65           & 0.80           & 0.78           & \textbf{0.90}           & 0.86           & 0.81           & 0.61           & 0.62           & 0.74           \\
Ours                 & \textbf{0.98}  & \textbf{0.78}  & \textbf{0.92}  & \textbf{0.90}  & \textbf{0.88}  & 0.87           & 0.83           & \textbf{0.86}  & \textbf{0.72}  & \textbf{0.88}  & \textbf{0.86}  \\ \hline
\multicolumn{12}{c}{PSNR (dB)}                                                                                                                                                                                       \\
\multicolumn{1}{l}{} & butterfly      & parrot         & house          & tulip          & bus            & river          & sea            & flower1        & flower2        & temple         & \textbf{ave.}        \\ \hline
Reinhard et al. \cite{reinhard2001color}      & 11.48          & 11.10           & 10.72           & 12.04           & 12.43           & 14.58           & 13.18           & 16.05  & 12.48           & 11.35          & 12.54           \\
Xiao et al. \cite{Xiao2006ColorTI}          & 11.17           & 12.18          & 13.83           & 12.14           & 11.83           & 15.25           & 12.96           & 11.93        & 10.43           & 11.72           & 12.34           \\
Petit et al. \cite{pitie2007linear}          & 14.94           & 11.23           & 12.64           & 11.13           & 11.94         & 15.26  & 13.36            & 15.68          & 11.91    & 10.63        & 12.87           \\
Petit et al. \cite{pitie2007automated}        & 11.17           & 13.45          & 14.42           & 12.19           & 12.05           & 15.31     & 13.15           & 15.77           & 11.74           & 11.73          & 13.10           \\
Ferradans et al. \cite{ferradans2013regularized}    & 18.70           & 12.13          & 14.31           & 16.65           & 11.77           & 15.47           & 13.42           & 16.37           & 13.56           & 11.78           & 14.42           \\
Rabin \& Papadakis \cite{rabin2014non}  & 9.66           & 11.21            & 13.12           & 13.09           & 10.51           & 15.15           & 13.11          & 15.34           & 11.44           & 14.96           & 12.76           \\
Ours   & \textbf{28.66}  & \textbf{20.71}  & \textbf{21.84}  & \textbf{22.88}  & \textbf{15.36}  & \textbf{16.89}   & \textbf{13.74}           & \textbf{18.15}  & \textbf{14.64}  & \textbf{21.65}  & \textbf{19.45}  \\ \hline
\end{tabular}
\end{table*}

\begin{figure*}[ht]
\centering{
\setcounter{subfigure}{0}
\subfloat[Source Image\label{fig:house_a}]{\includegraphics[width=33mm,scale=0.5]{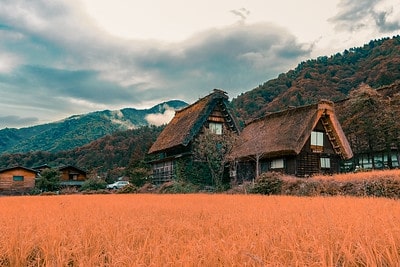}}\hspace{2.5em}%
\subfloat[Example Image\label{fig:house_b}]{\includegraphics[width=23mm,scale=0.5]{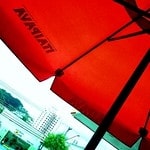}}\hspace{2.5em}%
\subfloat[Reinhard et al. \cite{reinhard2001color}\label{fig:house_c}]{\includegraphics[width=33mm,scale=0.5]{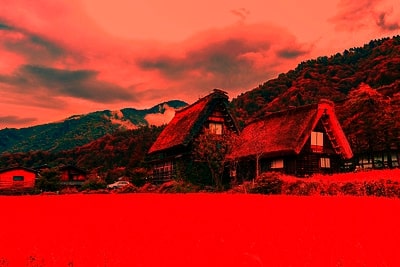}}
\hspace{1.0em}%
\subfloat[Xiao et al. \cite{Xiao2006ColorTI}\label{fig:house_d}]{\includegraphics[width=33mm,scale=0.5]{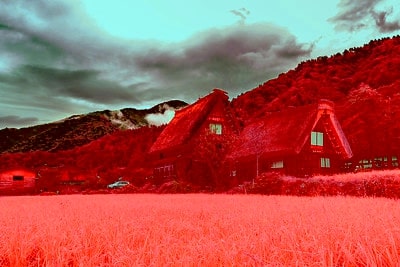}}}

{\subfloat[Petit et al. \cite{pitie2007linear}\label{fig:house_e}]{\includegraphics[width=33mm,scale=0.5]{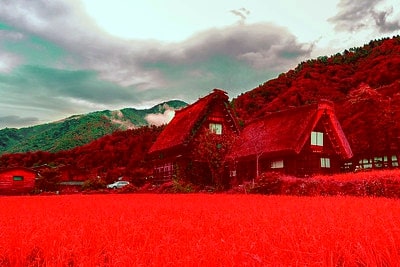}}\hspace{1.0em}%
\subfloat[Petit et al. \cite{pitie2007automated}\label{fig:house_f}]{\includegraphics[width=33mm,scale=0.5]{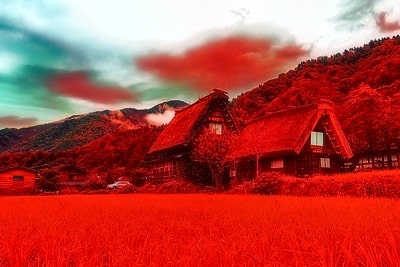}}\hspace{1.0em}%
\subfloat[Ferradans et al. \cite{ferradans2013regularized}\label{fig:house_g}]{\includegraphics[width=33mm,scale=0.5]{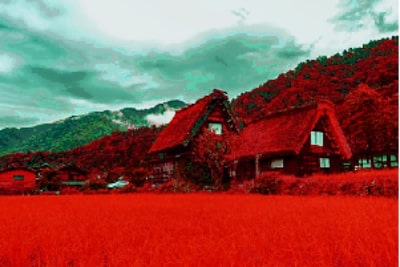}}\hspace{1.0em}%
\subfloat[Rabin \& Papadakis \cite{rabin2014non}\label{fig:house_h}]{\includegraphics[width=33mm,scale=0.5]{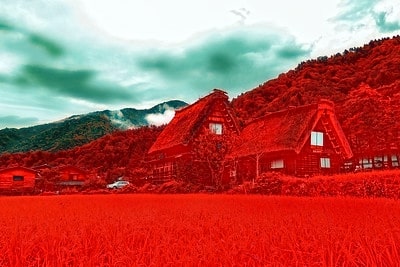}}}

{\subfloat[Ours ($q=5$) \label{fig:house_i}]{\includegraphics[width=33mm,scale=0.5]{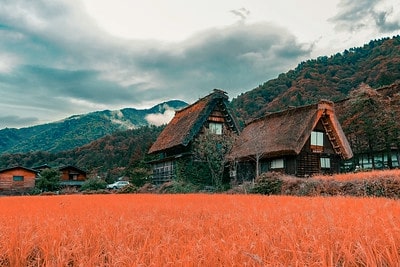}}\hspace{1.0em}%
\subfloat[Ours ($q=10$)\label{fig:house_j}]{\includegraphics[width=33mm,scale=0.5]{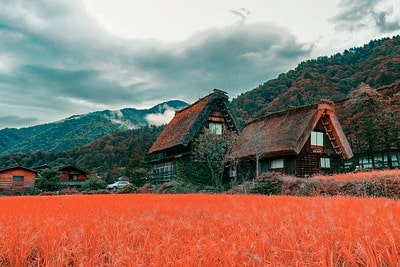}}\hspace{1.0em}%
\subfloat[Ours ($q=20$)\label{fig:house_k}]{\includegraphics[width=33mm,scale=0.5]{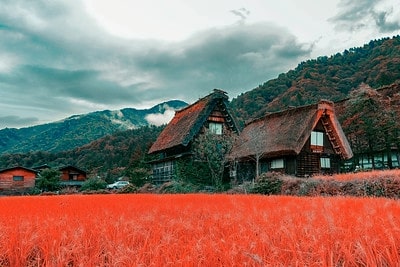}}\hspace{1.0em}%
\subfloat[Ours (final)\label{fig:house_l}]{\includegraphics[width=33mm,scale=0.5]{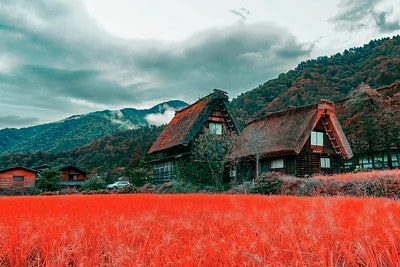}}}
\\[0.1ex]

\caption{Comparative results of \textit{house} in the dataset. The last row shows our results. }
\label{fig:house}
\end{figure*}

\begin{figure*}[t]
\centering
\setcounter{subfigure}{0}
{\subfloat[Source Image\label{fig:parrot_a}]{\includegraphics[width=33mm,scale=0.5]{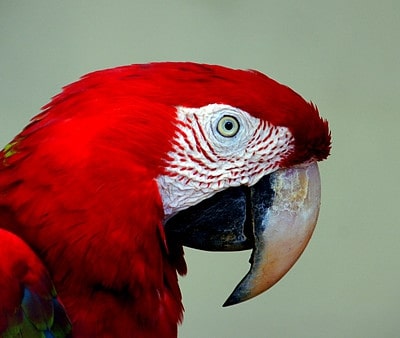}}\hspace{1.0em}%
\subfloat[Example Image\label{fig:parrot_b}]{\includegraphics[width=33mm,scale=0.5]{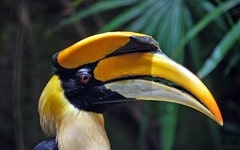}}\hspace{1.0em}%
\subfloat[Reinhard et al. \cite{reinhard2001color}\label{fig:parrot_c}]{\includegraphics[width=33mm,scale=0.5]{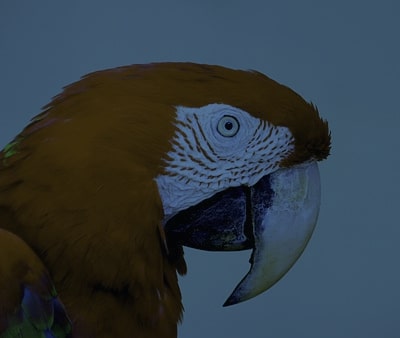}}
\hspace{1.0em}%
\subfloat[Xiao et al. \cite{Xiao2006ColorTI}\label{fig:parrot_d}]{\includegraphics[width=33mm,scale=0.5]{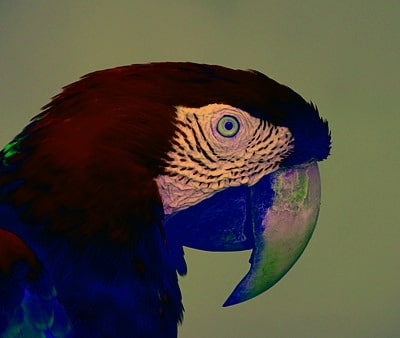}}
}

{\subfloat[Petit et al. \cite{pitie2007linear}\label{fig:parrot_e}]{\includegraphics[width=33mm,scale=0.5]{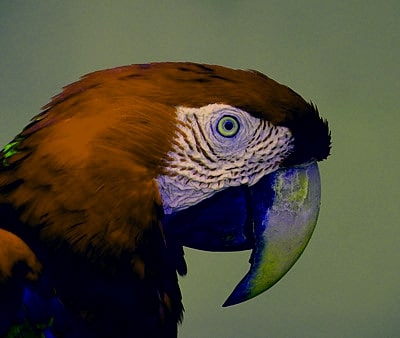}}\hspace{1.0em}%
\subfloat[Petit et al. \cite{pitie2007automated}\label{fig:parrot_f}]{\includegraphics[width=33mm,scale=0.5]{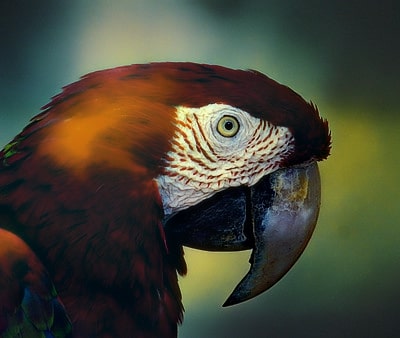}}\hspace{1.0em}%
\subfloat[Ferradans et al. \cite{ferradans2013regularized}\label{fig:parrot_g}]{\includegraphics[width=33mm,scale=0.5]{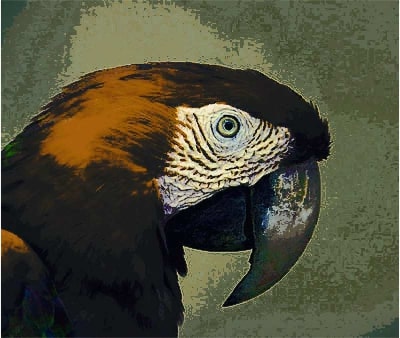}}\hspace{1.0em}%
\subfloat[Rabin \& Papadakis \cite{rabin2014non}\label{fig:parrot_h}]{\includegraphics[width=33mm,scale=0.5]{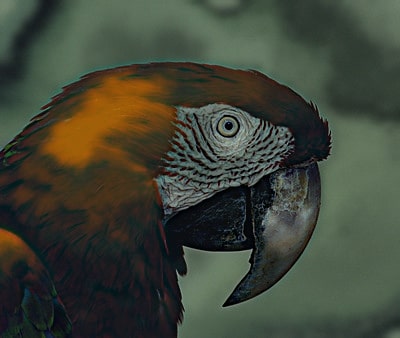}}}

{\subfloat[Ours ($q=5$)\label{fig:parrot_i}]{\includegraphics[width=33mm,scale=0.5]{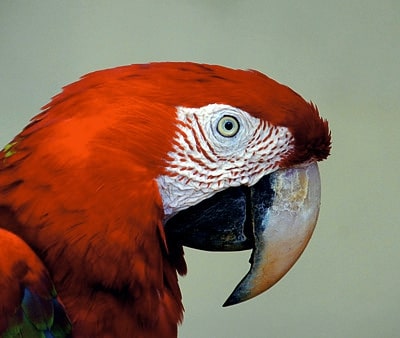}}\hspace{1.0em}%
\subfloat[Ours ($q=10$)\label{fig:parrot_j}]{\includegraphics[width=33mm,scale=0.5]{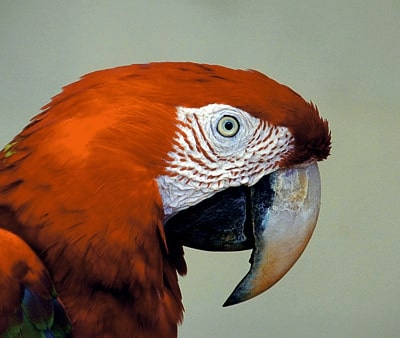}}\hspace{1.0em}%
\subfloat[Ours ($q=20$)\label{fig:parrot_k}]{\includegraphics[width=33mm,scale=0.5]{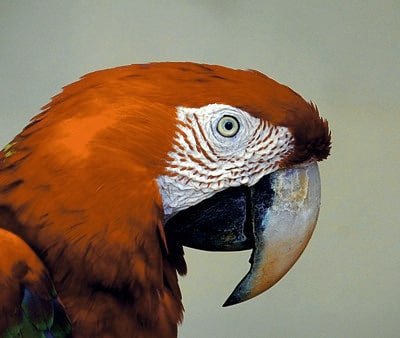}}\hspace{1.0em}%
\subfloat[Ours (final)\label{fig:parrot_l}]{\includegraphics[width=33mm,scale=0.5]{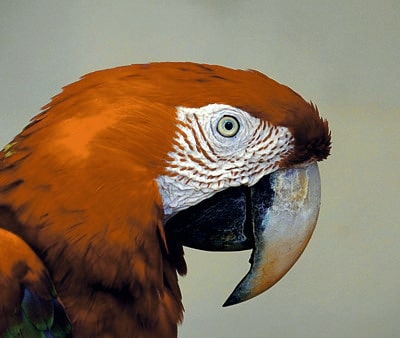}}}
\\[0.1ex]

\caption{Comparative results of \textit{parrot} in the dataset. The last row shows our results. }
\label{fig:parrot}
\end{figure*}

\begin{figure}[t]
\begin{center}
\includegraphics[width=0.7\linewidth]{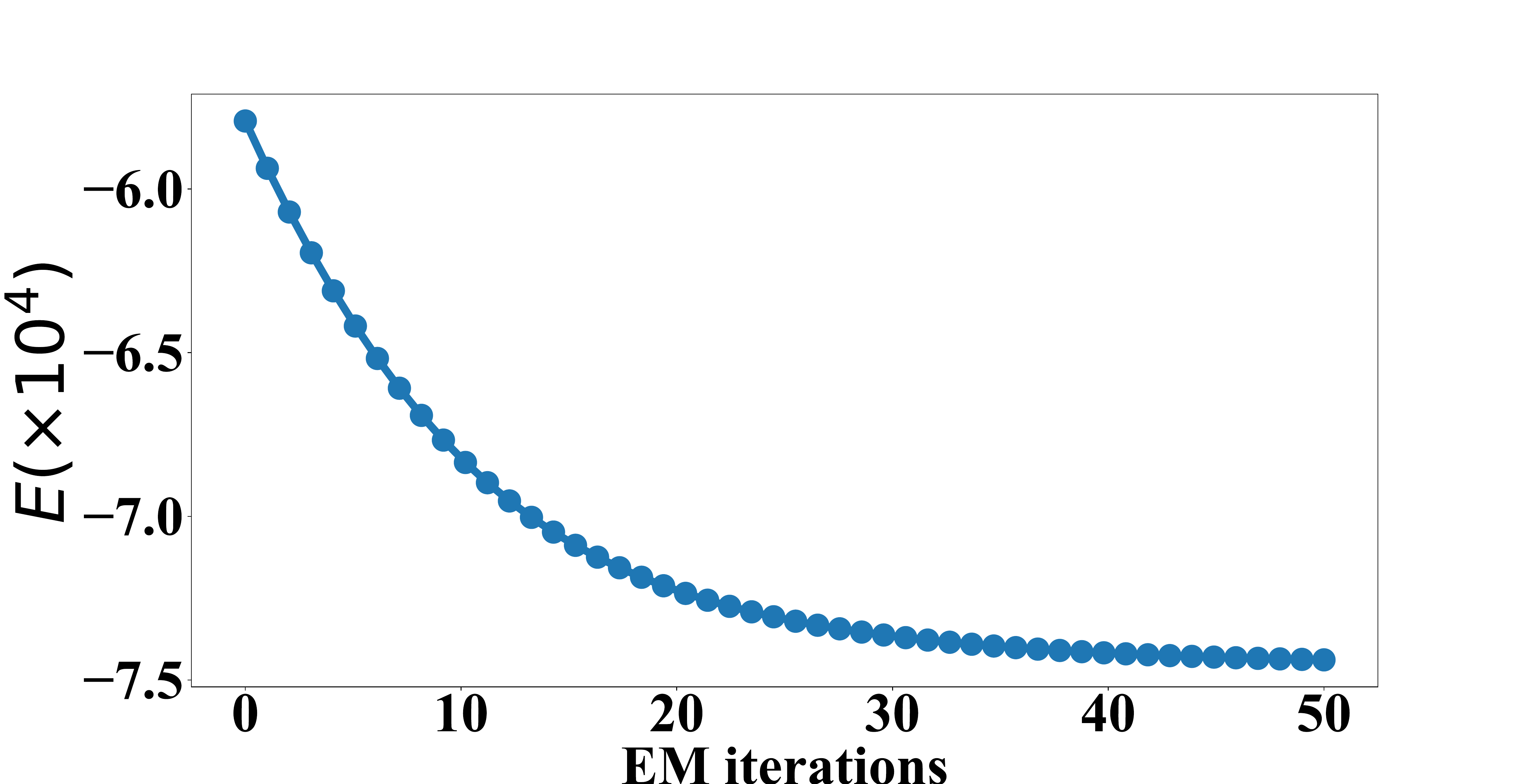}
\end{center}
\caption{An example for changes of Eq. \ref{eq:eq2} with increasing EM iterations. 
}
\label{fig:convergence}
\end{figure}

\subsection{Parameter Setting}
We set the termination criteria as the maximum number of iterations $q_{max} = 50$ being reached. Updating the $M \times K$ probability matrix $\{p_{mk}\}$ will lead to a vast computational cost. To alleviate this issue, the nearest neighbors in $Y$ of each $x_m$ are adopted instead of the whole $Y$. The search of nearest neighbors is conducted only once, since the change is negligible among iterations. The kernel size of Laplacian is empirically set to $5\times5$. All the parameter values are listed in Alg. \ref{alg:algorithm1}. To show the robustness of our method, we empirically fix most parameters and only tune $\mu$ in a very small range ($[0.001,0.005]$).

\begin{figure*}[t]
\centering
\setcounter{subfigure}{0}
{\subfloat[Source Image]{\includegraphics[width=33mm,scale=0.5]{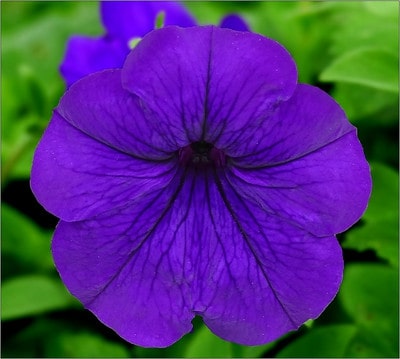}}\hspace{1.0em}%
\subfloat[Example Image]{\includegraphics[width=33mm,scale=0.5]{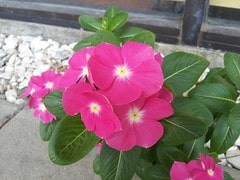}}\hspace{1.0em}%
\subfloat[Reinhard et al. \cite{reinhard2001color}]{\includegraphics[width=33mm,scale=0.5]{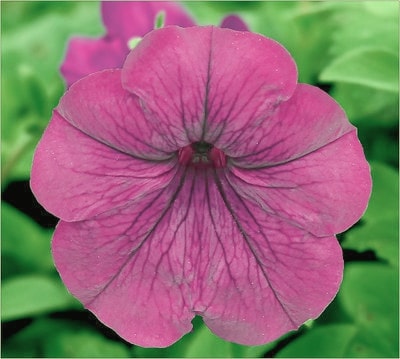}}
\hspace{1.0em}%
\subfloat[Xiao et al. \cite{Xiao2006ColorTI}]{\includegraphics[width=33mm,scale=0.5]{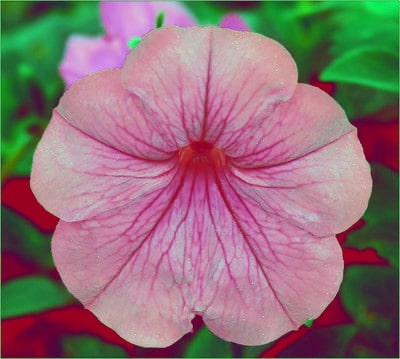}}
}

{\subfloat[Petit et al. \cite{pitie2007linear}]{\includegraphics[width=33mm,scale=0.5]{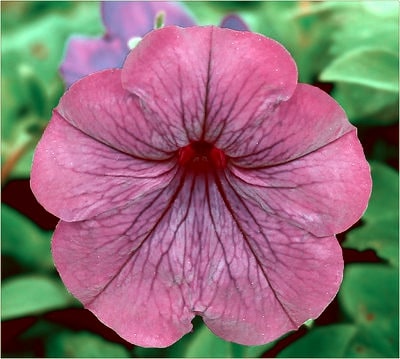}}\hspace{1.0em}%
\subfloat[Petit et al. \cite{pitie2007automated}]{\includegraphics[width=33mm,scale=0.5]{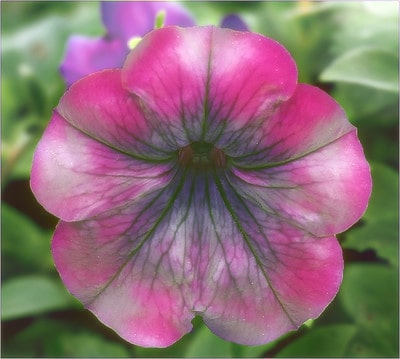}}\hspace{1.0em}%
\subfloat[Ferradans et al. \cite{ferradans2013regularized}]{\includegraphics[width=33mm,scale=0.5]{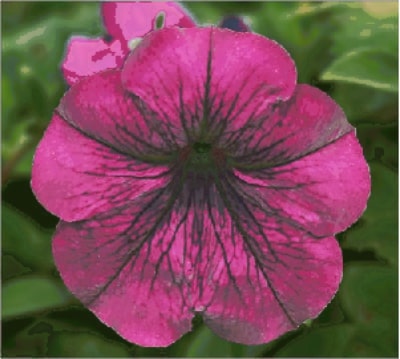}}\hspace{1.0em}%
\subfloat[Rabin \& Papadakis \cite{rabin2014non}]{\includegraphics[width=33mm,scale=0.5]{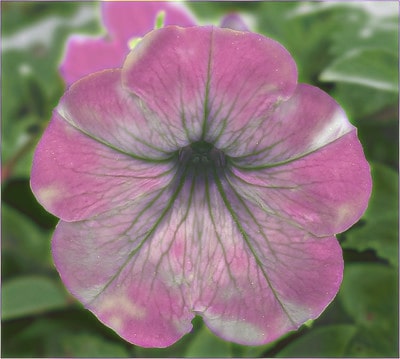}}}

{\subfloat[Ours ($q=5$)]{\includegraphics[width=33mm,scale=0.5]{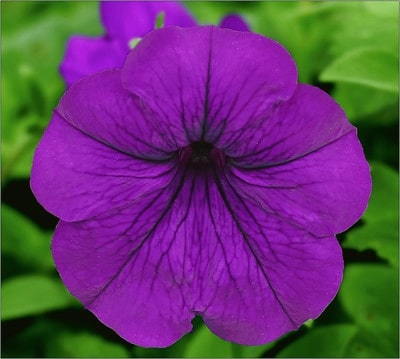}}\hspace{1.0em}%
\subfloat[Ours ($q=10$)]{\includegraphics[width=33mm,scale=0.5]{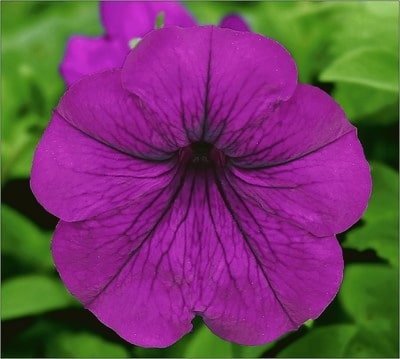}}\hspace{1.0em}%
\subfloat[Ours ($q=20$)]{\includegraphics[width=33mm,scale=0.5]{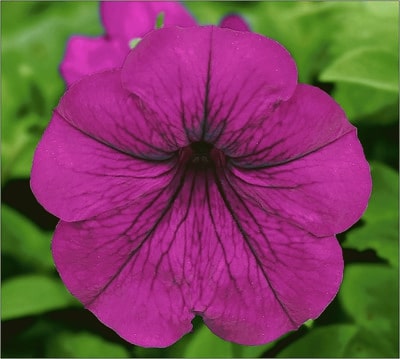}}\hspace{1.0em}%
\subfloat[Ours (final)]{\includegraphics[width=33mm,scale=0.5]{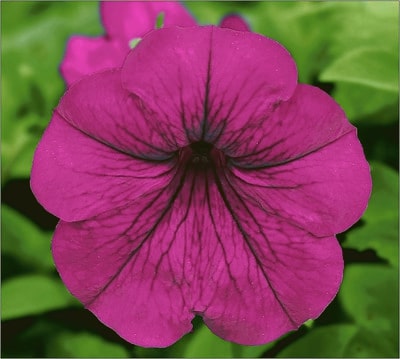}}}
\\[0.1ex]

\caption{Comparative results of \textit{flower2} in the dataset. The last row shows our results. }
\label{fig:flower2}
\end{figure*}

\begin{figure*}[t]
\centering
\setcounter{subfigure}{0}
\subfloat[Source Image]{\includegraphics[width=0.19\linewidth]{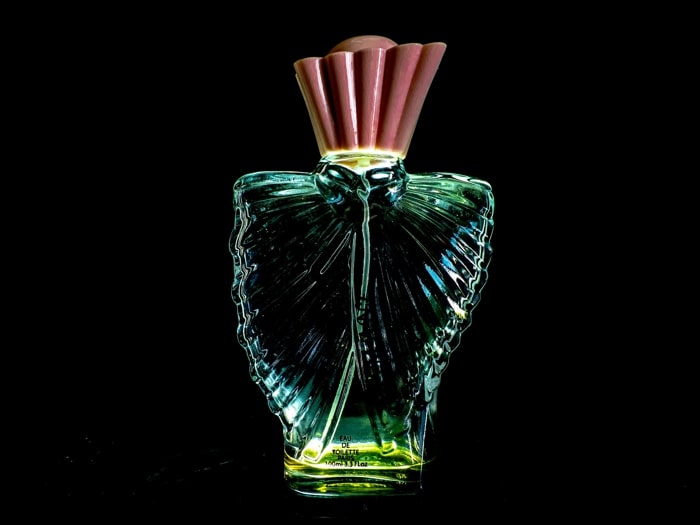}}\hspace{1.0em}%
\subfloat[Example Image]{\includegraphics[width=0.19\linewidth]{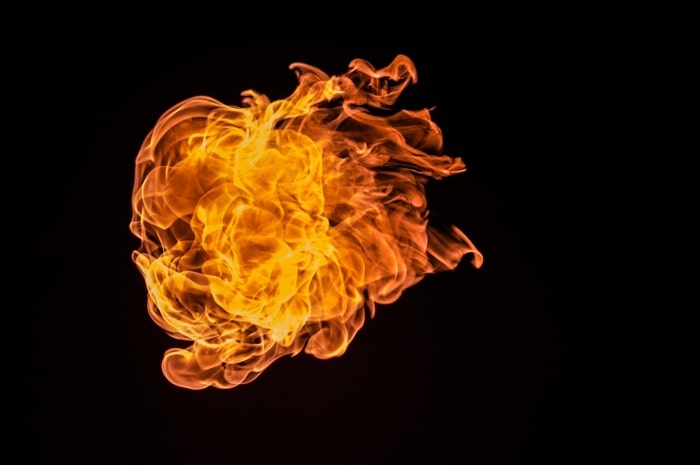}}\hspace{1.0em}
\subfloat[Luan et al. \cite{luan2017deep}]{\includegraphics[width=0.19\linewidth]{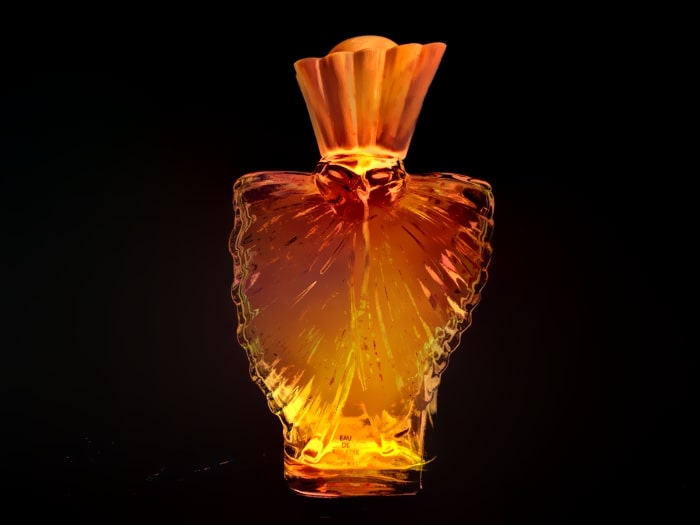}}\hspace{1.0em}
\subfloat[He et al. \cite{he2019progressive}]{\includegraphics[width=0.19\linewidth]{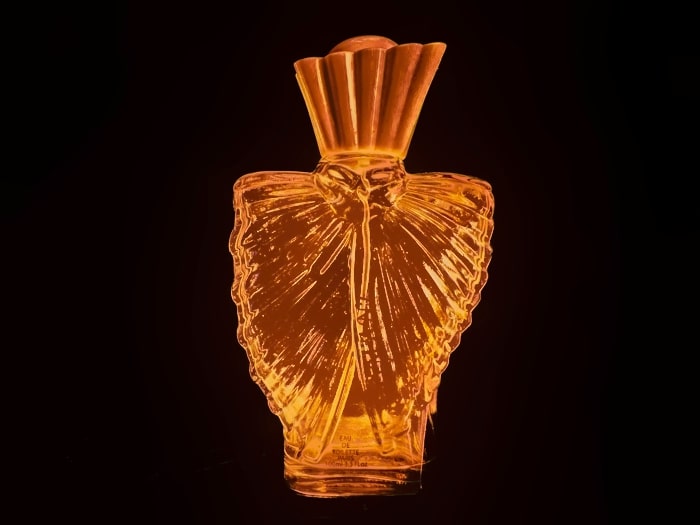}}

\subfloat[Ours ($q=5$)]{\includegraphics[width=0.19\linewidth]{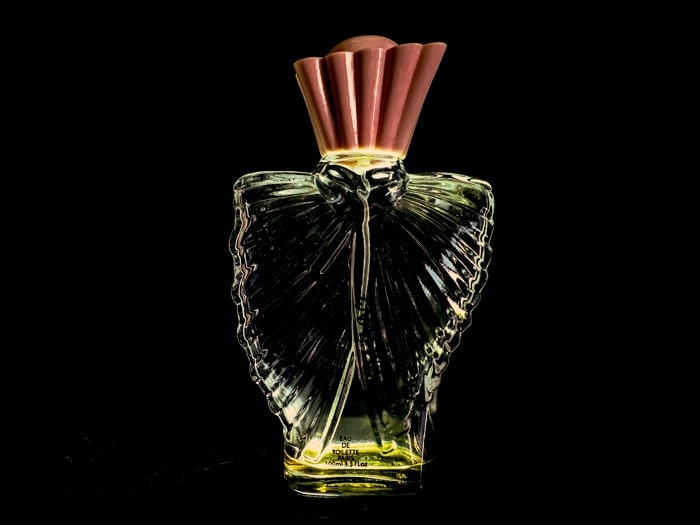}}\hspace{1.0em}
\subfloat[Ours ($q=10$)]{\includegraphics[width=0.19\linewidth]{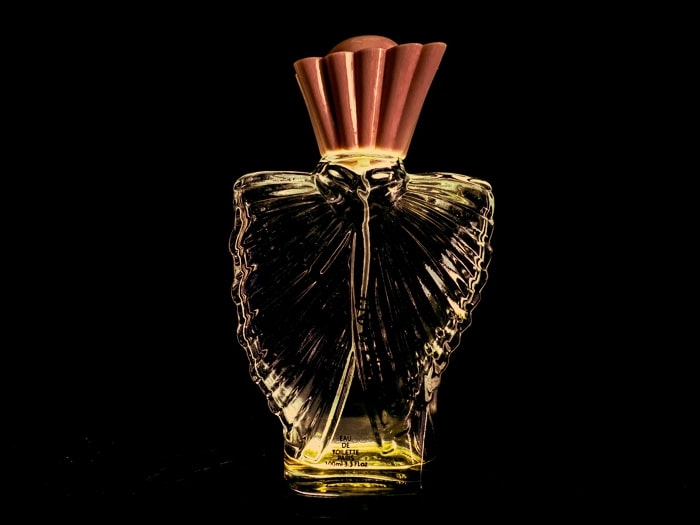}}\hspace{1.0em}
\subfloat[Ours ($q=20$)]{\includegraphics[width=0.19\linewidth]{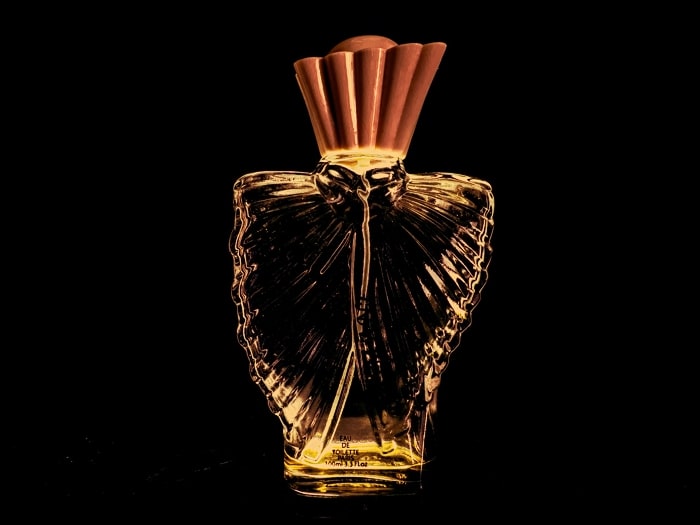}}\hspace{1.0em}
\subfloat[Ours (final)]{\includegraphics[width=0.19\linewidth]{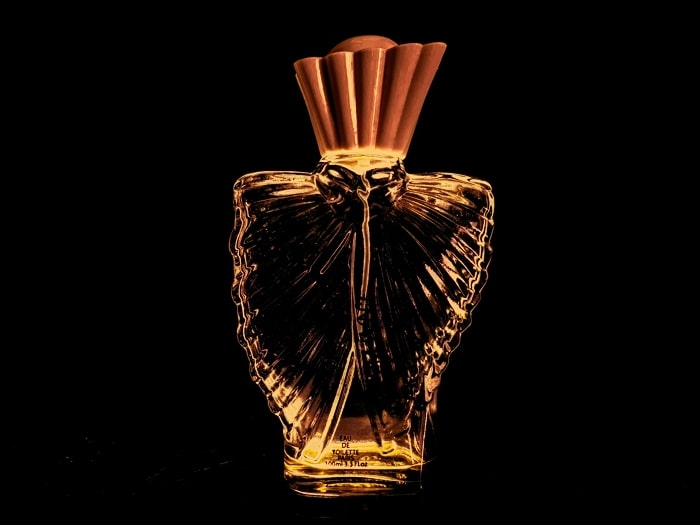}}

\caption{Comparative results against two deep leaning methods on \textit{perfume} (from \cite{luan2017deep}). (e) $\sim$ (h) show our results. The PSNR (dB)/SSIM of (c), (d) and (h) are 14.98/0.55, 15.35/0.61 and 23.17/0.94, respectively. }
\label{fig:perfume}
\end{figure*}

\begin{figure*}[t]
\centering
\setcounter{subfigure}{0}
\subfloat[Source Image]{\includegraphics[width=0.19\linewidth]{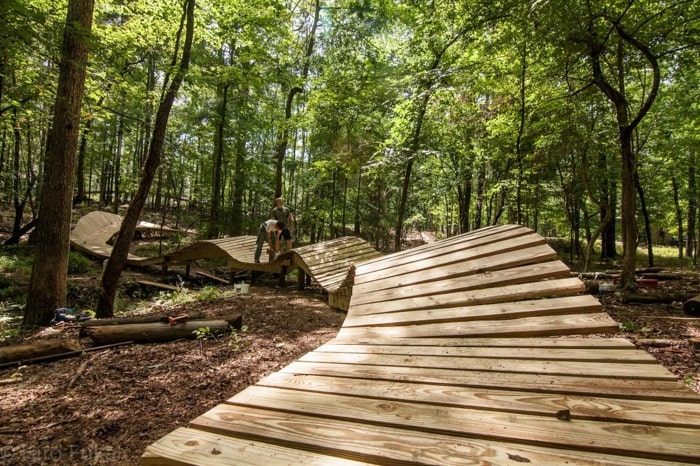}}\hspace{1.0em}
\subfloat[Example Image]{\includegraphics[width=0.19\linewidth]{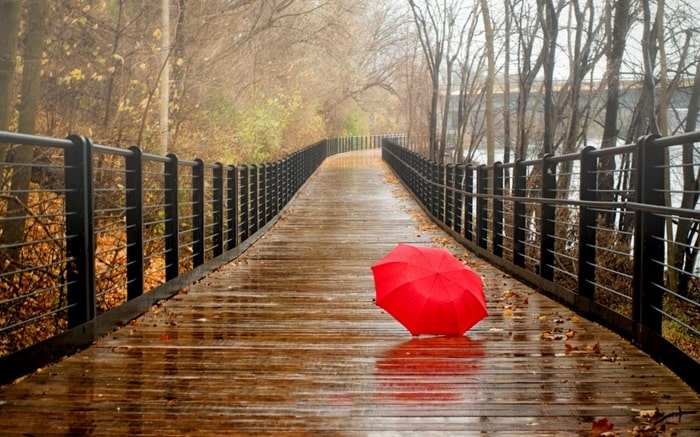}}\hspace{1.0em}
\subfloat[Luan et al. \cite{luan2017deep}]{\includegraphics[width=0.19\linewidth]{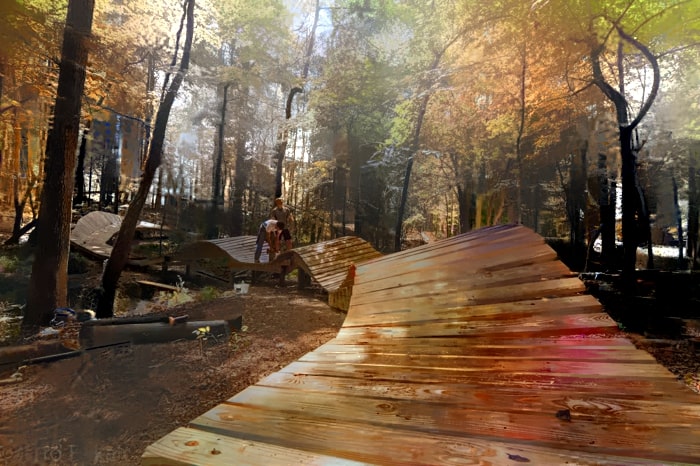}}\hspace{1.0em}
\subfloat[He et al. \cite{he2019progressive}]{\includegraphics[width=0.19\linewidth]{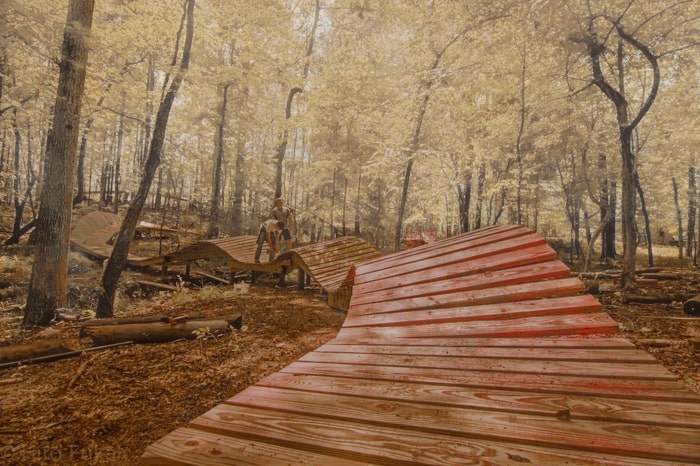}}

\subfloat[Ours ($q=5$)]{\includegraphics[width=0.19\linewidth]{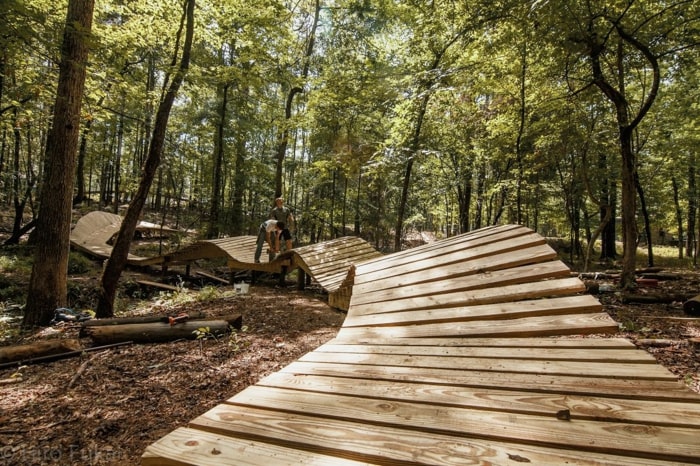}}\hspace{1.0em}
\subfloat[Ours ($q=10$)]{\includegraphics[width=0.19\linewidth]{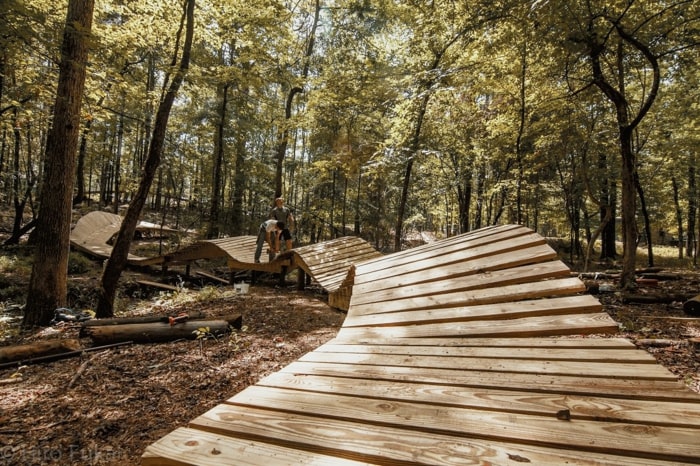}}\hspace{1.0em}
\subfloat[Ours ($q=20$)]{\includegraphics[width=0.19\linewidth]{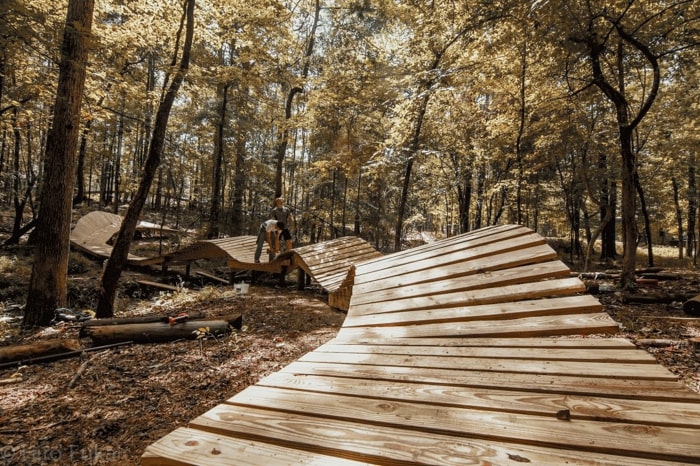}}\hspace{1.0em}
\subfloat[Ours (final)]{\includegraphics[width=0.19\linewidth]{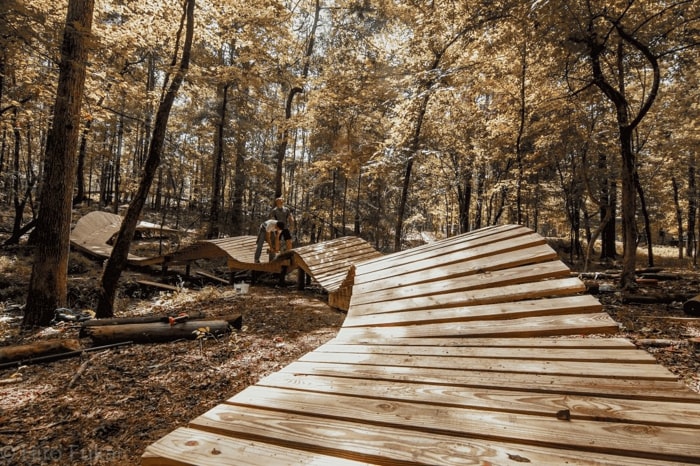}}

\caption{Comparative results against two deep leaning methods on \textit{road} (from \cite{luan2017deep}). (e) $\sim$ (h) show our results. The PSNR (dB)/SSIM of (c), (d) and (h) are 13.66/0.53, 13.07/0.68 and 24.89/0.96, respectively. }
\label{fig:road}
\end{figure*}

\begin{figure*}[t]
\centering
\setcounter{subfigure}{0}
\subfloat[Source Image]{\includegraphics[width=0.19\linewidth]{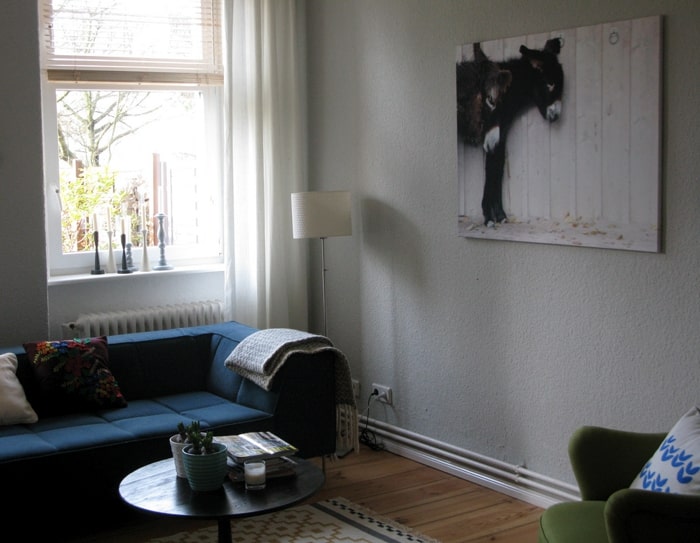}}\hspace{1.0em}
\subfloat[Example Image]{\includegraphics[width=0.19\linewidth]{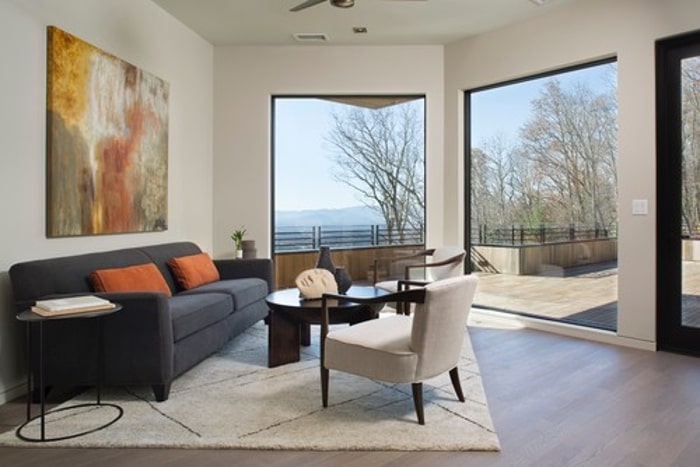}}\hspace{1.0em}
\subfloat[Luan et al. \cite{luan2017deep}]{\includegraphics[width=0.19\linewidth]{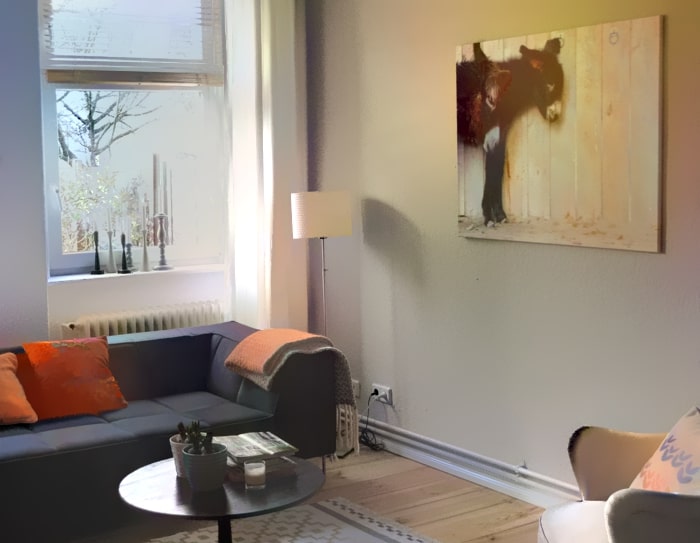}}\hspace{1.0em}
\subfloat[He et al. \cite{he2019progressive}]{\includegraphics[width=0.19\linewidth]{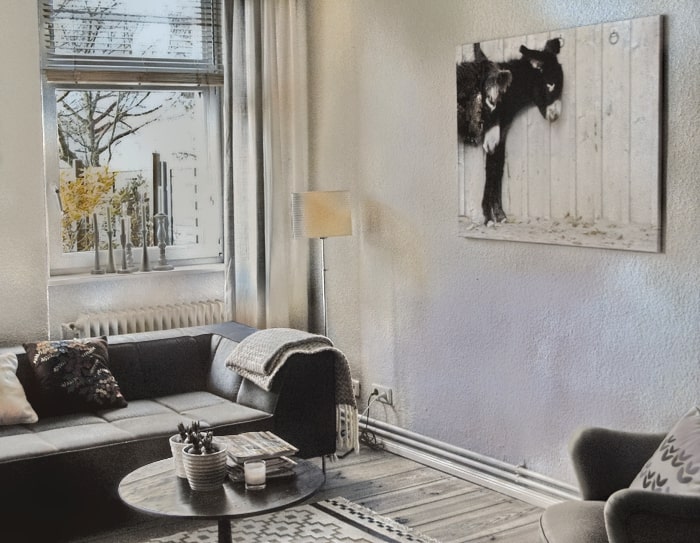}}

\subfloat[Ours ($q=5$)]{\includegraphics[width=0.19\linewidth]{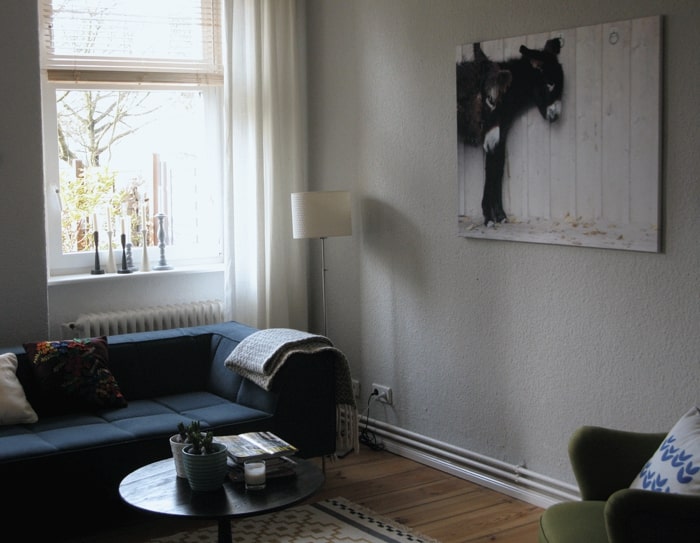}}\hspace{1.0em}
\subfloat[Ours ($q=10$)]{\includegraphics[width=0.19\linewidth]{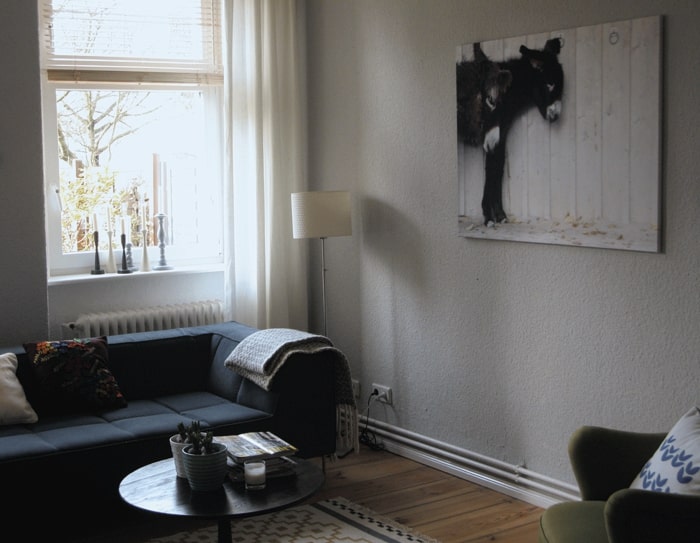}}\hspace{1.0em}
\subfloat[Ours ($q=20$)]{\includegraphics[width=0.19\linewidth]{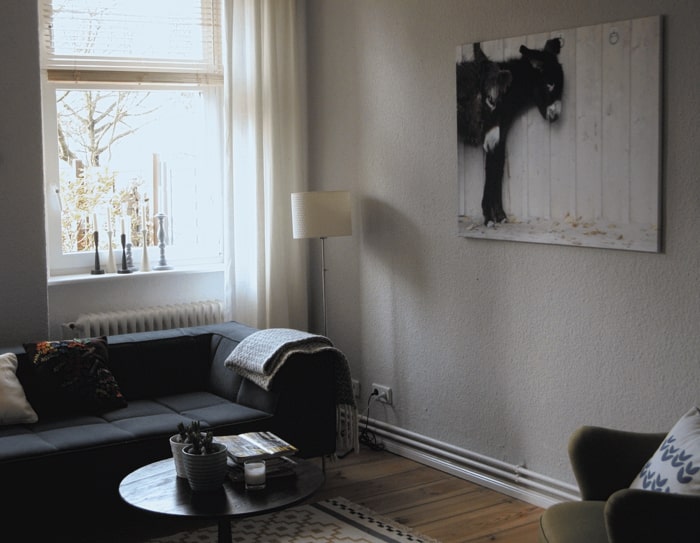}}\hspace{1.0em}
\subfloat[Ours (final)]{\includegraphics[width=0.19\linewidth]{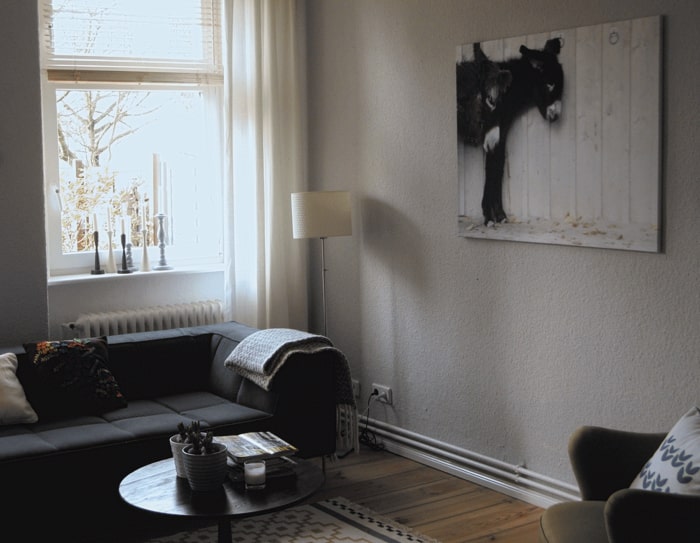}}

\caption{Comparative results against two deep leaning methods on \textit{home} (from \cite{luan2017deep}). (e) $\sim$ (h) show our results.}
\label{fig:home}
\end{figure*}

\subsection{Visual and Quantitative Results}
\label{sec:results}
We evaluate the performance of our approach against state-of-the-art 
color transfer methods with released source codes, both visually and quantitatively. 
In particular, we compare our method with the baseline approaches \cite{reinhard2001color,Xiao2006ColorTI}, two color distribution based methods \cite{pitie2007automated, pitie2007linear}, as well as two optimal transportation based techniques \cite{ferradans2013regularized, rabin2014non}. We also compare our method with two recent deep learning based techniques \cite{luan2017deep,he2019progressive} at a later stage (last paragraph in Section \ref{sec:results}). 

\textbf{Quantitative comparisons.} As for the quantitative evaluations, we adopt two major metrics, SSIM (Structural SIMilarity) \cite{wang2004image} and PSNR (Peak Signal to Noise Ratio), as suggested by the authors of previous research \cite{frigo2014optimal,hwang2014color}. The results are summarized in Tab. \ref{tab:metricsEvaluation}, which are computed between the output images and their corresponding source images. Note that although our method can generate a series of color transfer results, we simply use the result in the final iteration for all comparisons. 
SSIM suggests the degree of artifacts caused by the methods. As can be observed in Tab. \ref{tab:metricsEvaluation}, our method outperforms other methods on average. This well verifies the effectiveness of our method, and indicates that less damage is brought to the structure of the source image while transferring color. PSNR, which describes the mean squared error between two images, 
is taken as a secondary metric used in this work. 
It can be observed in Tab. \ref{tab:metricsEvaluation} that our method generates higher PSNR results, revealing that our color transfer method induces less totally new information (e.g., Fig. \ref{fig:perfume}). 

Fig. \ref{fig:convergence} shows an example for the changes of Eq. \ref{eq:eq2} with increasing EM iterations.
It can be seen that the negative loglikelihood declines drastically in the first $20$ iterations, and tend to be steady (i.e., converged) with increasing iterations.

\textbf{Visual comparisons.} We then contrast the visual results by taking a closer look at the test images \textit{house}, \textit{parrot} and \textit{flower2} in our test set\footnote{The data will be made publicly available.}. The comparison results are shown in Fig. \ref{fig:house}, Fig. \ref{fig:parrot} and Fig. \ref{fig:flower2}, respectively. Since the color transfer by \cite{reinhard2001color} is realized by stretching the input color histogram, it tends to produce global over-saturated color (e.g., Fig. \ref{fig:house}(c) and Fig. \ref{fig:parrot}(c)). An improved work that further considers color correlation \cite{Xiao2006ColorTI} alleviates this issue to some extent, but the results still look unnatural, as shown in Fig. \ref{fig:house}(d), Fig. \ref{fig:parrot}(d) and Fig. \ref{fig:flower2}(d). \cite{pitie2007linear, pitie2007automated} aim to find an appropriate color mapping, linearly and non-linearly. However, from the results we can see that the red color in the example image is incorrectly mapped to the forest in Fig. \ref{fig:house}(e), though there exist green pixels in the example image. \cite{ferradans2013regularized, rabin2014non} build the color transfer maps based on an estimated optimal transportation to connect two input images. \cite{ferradans2013regularized} puts emphasis on reducing the artifacts by adding transport regularization, while artifacts caused by \gu{two highly different distributions of source and example in shape during transportation} can still be observed (e.g., Fig. \ref{fig:parrot}(g)). \cite{rabin2014non} adopts discrete optimal transport in color transfer with extra attention to color and spatial relationship. However, it can be seen from Fig. \ref{fig:parrot}(h) and \ref{fig:flower2}(h) that it causes over-smoothness in local regions with similar colors, which leads to artifacts near boundaries and loss of sharp features. Additionally, these methods can only produce a single result for each input pair of images, inducing no other choices for users.

Our method, on the other hand, has the capacity to generate more visually pleasant results with less artifacts, and preserve sharp features better, shown in Fig. \ref{fig:regularizationComparison}. 
Also, it is interesting that our approach can create a variety of color transfer results with increasing the number of iterations, which provides flexibility to users in the real-world applications, as illustrated in Fig. \ref{fig:house}(i-l), Fig. \ref{fig:parrot}(i-l) and Fig. \ref{fig:flower2}(i-l). These results seem natural and show gradual changes of the color transfer, without requiring to build any color correspondences.

\textbf{Comparison with deep learning based methods.}
Given that deep learning based methods require a pre-defined input image size, we thus choose image pairs from two recent deep learning methods \cite{luan2017deep,he2019progressive} and compare our method with them. In essence, \cite{luan2017deep} is a style transfer method and \cite{he2019progressive} is a color transfer technique. We show the comparisons in Fig. \ref{fig:perfume} $\sim$ \ref{fig:home}. It can be observed in Fig. \ref{fig:perfume} that our method produces more natural results than the deep learning methods \cite{luan2017deep,he2019progressive}, in terms of preserving the original structure without introducing extra content in the result. In Fig. \ref{fig:road}, both \cite{luan2017deep} and \cite{he2019progressive} are affected greatly by the red umbrella in Fig. \ref{fig:road}(b), and their tree shadows (left area in the image) seem to be corrupted. Also, it seems that the result by \cite{luan2017deep} involves artifacts such as ``halos'' which blur the background tree area. Continuous color transfer results can be generated by our method. In Fig. \ref{fig:home}, the result by our method resembles the source image, because the two input images are highly similar in tone of color. This is a failure case for our method. By contrast, in this case, the two deep learning methods \cite{luan2017deep,he2019progressive} can perform transfer successfully in Fig. \ref{fig:home}(c,d). This is because these two deep learning methods could learn high-level features which somewhat enables the semantic relationship between two input images. Still, over-smoothness (e.g., the window in Fig. \ref{fig:home}(c)) and unnatural places (e.g., the wall in Fig. \ref{fig:home}(d)) can be seen.

\section{Conclusion}
In this paper, we have proposed a novel approach for pixel-wise color transfer. Given a source image and an example image as input, it enables the color style of the source image to approximate the example image with an iterative optimization. Our method can generate multiple transfer results and does not require the two input images to be the same resolution. Also, we introduced a regularization term, which allows gradient information preservation and smooth color transition. Experiments show that our method generally outperforms current methods, both qualitatively and quantitatively. The major limitation of our method is that it can hardly generate desired results for two input images with highly similar color tone (see Fig. \ref{fig:home}). In the future, we would like to explore semantic relationships between two images and incorporate it into our framework.

\ifCLASSOPTIONcaptionsoff
  \newpage
\fi



%
\bibliographystyle{abbrv}
\bibliography{reference}

%






\end{document}